\documentclass[10pt,twocolumn,letterpaper]{article}

\usepackage{iccv}
\usepackage{times}
\usepackage{epsfig}
\usepackage{graphicx}
\usepackage{amsmath}
\usepackage{amssymb}

\usepackage{color}
\usepackage{multirow}
\usepackage{tabularx}
\usepackage{booktabs}
\usepackage{makecell}
\usepackage[table,dvipsnames]{xcolor}
\usepackage[ruled,linesnumbered]{algorithm2e}
\usepackage{pifont}
\usepackage[accsupp]{axessibility}

\usepackage{amsmath,amsfonts,bm}

\def\eqref#1{equation~\ref{#1}}

\def\1{\bm{1}}

\def\vmu{{\bm{\mu}}}

\def\va{{\bm{a}}}

\def\mB{{\bm{B}}}

\def\mD{{\bm{D}}}

\def\mF{{\bm{F}}}

\def\mI{{\bm{I}}}

\def\mM{{\bm{M}}}

\def\mQ{{\bm{Q}}}

\def\mT{{\bm{T}}}

\def\mW{{\bm{W}}}

\DeclareMathAlphabet{\mathsfit}{\encodingdefault}{\sfdefault}{m}{sl}
\SetMathAlphabet{\mathsfit}{bold}{\encodingdefault}{\sfdefault}{bx}{n}

\def\sR{{\mathbb{R}}}
\def\sS{{\mathbb{S}}}

\def\sZ{{\mathbb{Z}}}

\def\cA{{{A}}}
\def\cB{{{B}}}

\definecolor{aliceblue}{rgb}{0.94, 0.97, 1.0}
\definecolor{citecolor}{HTML}{0071BC}
\definecolor{linkcolor}{HTML}{ED1C24}
\definecolor{darkgreen}{HTML}{539165}

\makeatletter
\newcommand{\thickhline}{%
 \noalign {\ifnum 0=`}\fi \hrule height 1pt
 \futurelet \reserved@a \@xhline
}
\makeatother

\newcommand{\increase}[1]{{
  \fontsize{6.6pt}{0.5em}\selectfont({\color{darkgreen}{$\uparrow$~\textbf{#1}}})
}}
\newcommand{\decrease}[1]{{
  \fontsize{6.6pt}{0.5em}\selectfont({\color{purple}{$\downarrow$~\textbf{#1}}})
}}

\usepackage[pagebackref, breaklinks=true, colorlinks, citecolor=NavyBlue, linkcolor=linkcolor, bookmarks=false]{hyperref}

\usepackage[capitalize]{cleveref}
\crefname{section}{Sec.}{Secs.}
\Crefname{section}{Section}{Sections}
\Crefname{table}{Table}{Tables}
\crefname{table}{Tab.}{Tabs.}

\def\iccvPaperID{****} 

\iccvfinalcopy

\begin{document}

\title{Multi-granularity Interaction Simulation for Unsupervised \\ Interactive Segmentation}

\author{
Kehan Li$^{1,3}$ \quad Yian Zhao$^{5}$ \quad Zhennan Wang$^{2}$ \quad Zesen Cheng$^{1,3}$ \quad Peng Jin$^{1,3}$ \\ Xiangyang Ji$^{4}$ \quad Li Yuan$^{1,2,3}$ \quad Chang Liu$^{4}$\thanks{Corresponding author.} \quad Jie Chen$^{1,2,3}$$^*$ \and
$^{1}$ School of Electronic and Computer Engineering, Peking University, Shenzhen, China \\
$^{2}$ Peng Cheng Laboratory, Shenzhen, China \\
$^{3}$ AI for Science (AI4S)-Preferred Program, Peking University, Shenzhen, China \\
$^{4}$ Department of Automation and BNRist, Tsinghua University, Beijing, China \\
$^{5}$ Dalian University of Technology, Dalian, China
}

\maketitle
\ificcvfinal\thispagestyle{empty}\fi

\begin{abstract}

Interactive segmentation enables users to segment as needed by providing cues of objects, which introduces human-computer interaction for many fields, such as image editing and medical image analysis.
Typically, massive and expansive pixel-level annotations are spent to train deep models by object-oriented interactions with manually labeled object masks.
In this work, we reveal that informative interactions can be made by simulation with semantic-consistent yet diverse region exploration in an unsupervised paradigm.
Concretely, we introduce a \textbf{M}ulti-granularity \textbf{I}nteraction \textbf{S}imulation (\textbf{MIS}) approach to open up a promising direction for unsupervised interactive segmentation.
Drawing on the high-quality dense features produced by recent self-supervised models, we propose to gradually merge patches or regions with similar features to form more extensive regions and thus, every merged region serves as a semantic-meaningful multi-granularity proposal.
By randomly sampling these proposals and simulating possible interactions based on them, we provide meaningful interaction at multiple granularities to teach the model to understand interactions.
Our MIS significantly outperforms non-deep learning unsupervised methods and is even comparable with some previous deep-supervised methods without any annotation.

\end{abstract}

\section{Introduction}

\begin{figure}[t]
    \centering
    \includegraphics[width=\linewidth]{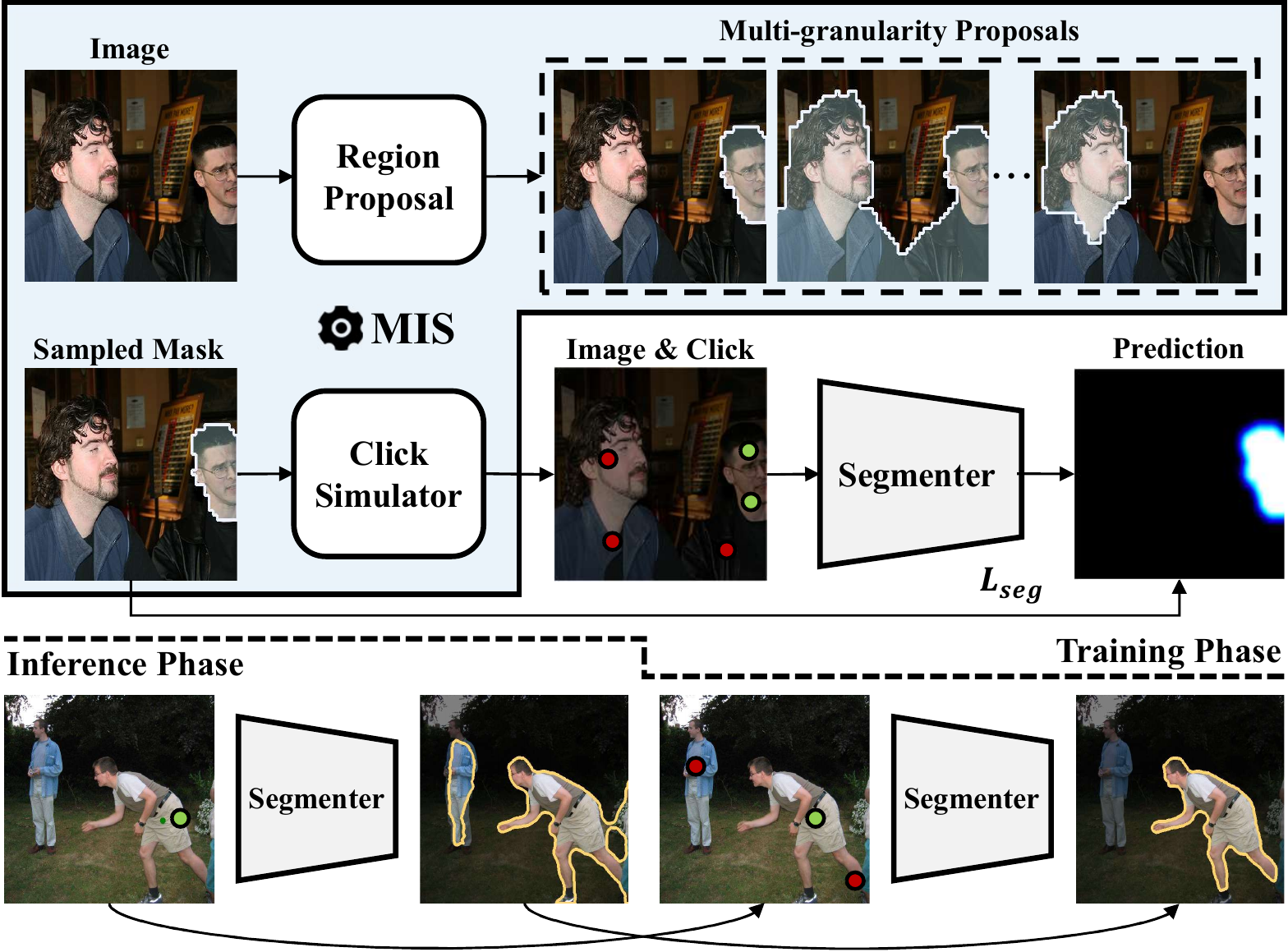}
    \caption{
        \textbf{Top:} To tackle the reliance on object annotations of previous interactive segmentation methods caused by interaction simulation, we propose \textbf{M}ulti-granularity \textbf{I}nteraction \textbf{S}imulation (\textbf{MIS}) that simulates interaction by meaningful regions at multiple granularities.
        \textbf{Down:} The model learns from MIS to successfully understand interactions and produce acceptable segmentation.
    }
    \label{fig:motivation}
\end{figure}

Interactive segmentation aims at obtaining high-quality object masks with limited user interaction, allowing users to segment objects as needed.
During the development of interactive segmentation, various interactive forms like bounding boxes~\cite{lempitsky2009image,rother2004grabcut,wu2014milcut}, scribbles~\cite{bai2014error,grady2006random,li2004lazy}, and clicks~\cite{xu2016deep,jang2019interactive,lin2020interactive,sofiiuk2020f,chen2021conditional,sofiiuk2022reviving,lin2022focuscut,chen2022focalclick} have been explored.
Among them, the click-based interaction becomes the mainstream due to its simplicity and well-established training and evaluation protocols.

\begin{figure*}[t]
    \centering
    \includegraphics[width=\textwidth]{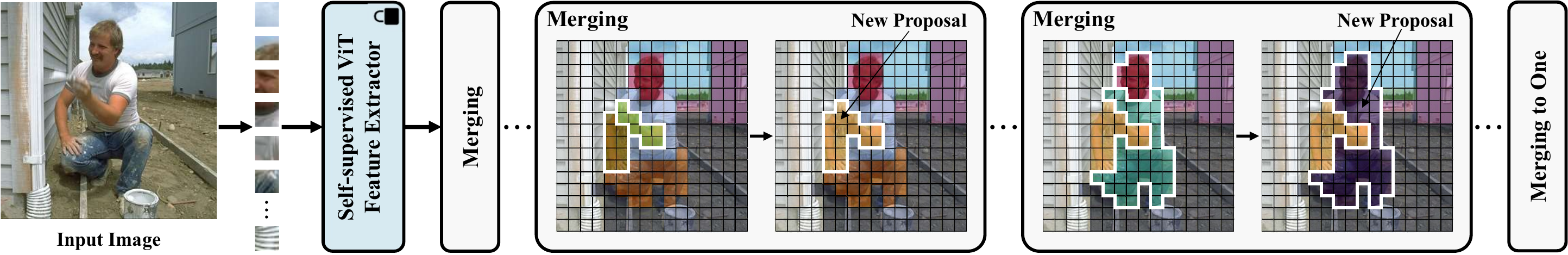}
    \caption{
        \textbf{Illustration of the multi-granularity region proposal generation.}
        For an input image, We first introduce a self-supervised ViT to produce semantic features for each patch of an image.
        We then gradually merge two similar patches or regions with similar features to perform a new region.
        All newly generated regions in the merging process constitute meaningful proposals at multiple granularities.
    }
    \label{fig:intuition}
\end{figure*}

State-of-the-art methods for click-based interactive segmentation are based on deep learning, following the basic paradigm proposed by Xu \emph{et al.}~\cite{xu2016deep}.
Specifically, these methods encode clicks to a distance map and adapt a semantic segmentation model (\emph{e.g.}, FCN~\cite{long2015fully}) to take the encoded click map as input, and then train the model with interaction-segmentation pairs.
Since it is impractical to collect interaction sequences from real users, previous methods adopt an interaction simulation strategy~\cite{xu2016deep,sofiiuk2022reviving} that randomly samples clicks based on object segmentation annotations.
By training with the simulated clicks and corresponding object masks, the model is able to correctly understand user needs and segment objects with a few clicks, significantly outperforming non-deep learning methods.
However, the requirement for a large number of expensive pixel-level annotations behind the powerful performance has been ignored by previous methods, which is derived from the object-level interaction simulation.

In this work, we aim at exploring an annotation-free alternation for interaction simulation to eliminate the reliance on object annotations in interactive segmentation, namely unsupervised interactive segmentation.
To achieve it, as shown in the top half of \Cref{fig:motivation}, our basic idea is to switch previous object-oriented interaction simulation to some semantic-consistent regions, which can be discovered unsupervisedly.
In this case, the model requires diverse examples to understand changeable interactions due to the lack of precise information about objects.

Therefore, we propose \textbf{M}ulti-granularity \textbf{I}nteraction \textbf{S}imulation (\textbf{MIS}) to achieve diversity.
To ensure the semantic consistency of a region, we first parse the semantic of patches in an image by a Vision Transformer (ViT)~\cite{dosovitskiy2020image} pre-trained unsupervisedly with DINO~\cite{caron2021emerging}.
As shown in \Cref{fig:intuition}, we then gradually merge two patches or regions with similar features to form a more extensive region for an image, until only one region remains.
In this process, the newly generated region in every step makes up semantic-meaningful multi-granularity proposals.
The MIS is finally achieved by randomly sampling these proposals and simulating possible interactions based on them when training the model.
Moreover, in order to help the model fight against the unsmoothness on the boundary of proposals caused by the feature extractor, we design a smoothness constraint that considers the low-level feature of pixels.
With the meaningful interaction simulated at multiple granularities and the smoothness constraint, the model learns to understand clicks by related regions and produce an acceptable segmentation with several user interactions when inference, while completely discarding object annotations, as shown in the bottom half of \Cref{fig:motivation}.

We take the standard protocol~\cite{xu2016deep} to evaluate our method on commonly used datasets including GrabCut~\cite{rother2004grabcut}, Berkeley~\cite{mcguinness2010comparative}, SBD~\cite{hariharan2011semantic}, and DAVIS~\cite{perazzi2016benchmark}.
Benefiting from the rich interaction examples from our MIS, the trained model achieves inspiring performance.
Specifically, it significantly outperforms non-deep learning methods under the premise of unsupervised setting, and even surpasses some early deep supervised methods.
In addition, it can also perform as an unsupervised pre-training to improve the performance when only limited annotations are available.
The surprising results support that the interactive segmentation task can be solved in a more label-efficient way.
Moreover, we believe the proposed method can reduce the labeling costs in segmentation tasks by training an interactive segmentation model unsupervisedly and using it to efficiently make task-specific annotations as needed.

\begin{figure*}[t]
    \centering
    \includegraphics[width=\textwidth]{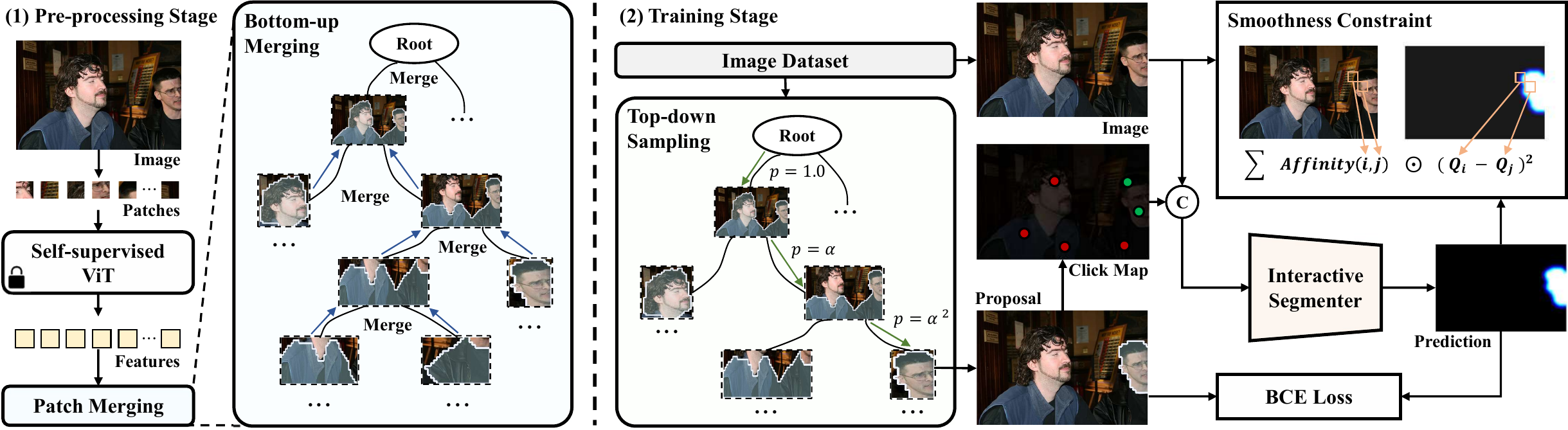}
    \caption{
        \textbf{Illustration of our unsupervised framework to train an interactive segmentation model.}
        For implementation, the MIS is composed of two stages.
        In the first pre-processing stage, we first parse the semantic of each patch in the image using a pre-trained ViT and take a gradual merging strategy to produce semantic-consistent region proposals at multiple granularities and save them efficiently with a tree structure.
        In the training stage, given an image, we randomly select proposals from the corresponding merging tree in a top-down manner and further simulate possible clicks based on them, thus providing the model informative interactions.
    }
    \label{fig:framework}
\end{figure*}

\section{Related Works}

\noindent \textbf{Interactive Image Segmentation.}
Interactive segmentation has been an active field for over two decades due to its wide applications.
Before the era of deep learning, the solutions are based on low-level features and optimization.
One of the best-known methods is proposed by Boykov and Jolly~\cite{boykov2001interactive}, who define a graph to describe user-provided hard constraints and pixel relationship, then solve interactive segmentation by graph cut with max-flow algorithm~\cite{boykov2004experimental}.
After that, Rother \emph{et al.}~\cite{rother2004grabcut} augment graph cut by more powerful color data modeling and an iterative strategy.
Other algorithms include random walk~\cite{grady2006random}, geodesic distance~\cite{bai2007geodesic}, and star-convexity~\cite{gulshan2010geodesic}.
Due to the dependence on low-level features, these methods stuck when processing complex images.
Xu \emph{el al.}~\cite{xu2016deep} are the first to introduce deep learning on this task by extending FCN~\cite{long2015fully} with click map and click sampling strategies.
Based on this pipeline, the researchers gradually increase the performance ceiling of deep learning methods by emphasizing the first click~\cite{lin2020interactive}, online optimization~\cite{jang2019interactive,sofiiuk2020f}, diffusing prediction~\cite{chen2021conditional}, edge-guided flow~\cite{hao2021edgeflow}, iterative click sampling~\cite{sofiiuk2022reviving}, powerful feature extractor~\cite{liu2022simpleclick}, and Gaussian process posterior~\cite{zhou2023interactive}.
On the other hand, some works~\cite{lin2022focuscut,chen2022focalclick} focus on efficiency and propose to segment on local crops to accelerate inference.
Although the deep learning approaches achieve uplifting performance and efficiency, they require large-scale pixel-level annotations to train, which are expensive and laborious to obtain.
Unlike previous works, we focus on the dependency on annotations and show that a fully unsupervised framework can also handle interactive segmentation task well.

\smallskip

\noindent \textbf{Unsupervised Image Segmentation.}
With the development of self-supervised and unsupervised learning, unsupervised methods for image segmentation task start to emerge.
This work is also related to some unsupervised image segmentation methods based on the idea that extracting segments from a pre-trained high-quality dense feature extractor, (\emph{e.g.}, DenseCL~\cite{wang2021dense}, DINO~\cite{caron2021emerging}).
For instance, Simeoni \emph{et al.}~\cite{simeoni2021localizing} proposed a series of hand-made rules to choose pixels belonging to the same object according to their feature similarity.
Wang \emph{et al.}~\cite{wang2022self} introduce normalized cuts~\cite{shi2000normalized} on the affinity graph constructed by pixel-level representations from DINO to divide the foreground and background of an image.
Hamilton \emph{et al.}~\cite{hamilton2022unsupervised} train a segmentation head by distilling the feature correspondences from DINO.
Melas \emph{et al.}~\cite{melas2022deep} adopt spectral decomposition on the affinity graph to discover meaningful parts in an image.
For instance segmentation, FreeSOLO~\cite{wang2022freesolo} design pseudo instance mask generation based on multi-scale feature correspondences from densely pre-trained models and train an instance segmentation model with these pseudo masks.
Although the masks produced by these methods can serve as pseudo labels to train an interactive segmentation model unsupervisedly, only ordinary results are observed through our experiments in \Cref{sec:results}, which demonstrate that these task-specific designs is not apposite for interactive segmentation task.
In contrast, we propose a novel strategy to discover semantic-consistent yet diverse regions to fit the interactive segmentation task.

\section{Method}

In this section, we introduce the overall framework to solve unsupervised interactive segmentation in detail.
Macroscopically, we simulate informative interactions from the proposed \textbf{M}ulti-granularity \textbf{I}nteraction \textbf{S}imulation (\textbf{MIS}) to train an interactive segmentation model. 
For implementation, the MIS is composed of two stages, as shown in \Cref{fig:framework}.
In the first pre-processing stage, we first parse the semantic of each patch in the image using a ViT~\cite{dosovitskiy2020image} trained self-supervisedly with DINO~\cite{caron2021emerging}, which has been proven to produce high-quality dense semantic features~\cite{simeoni2021localizing,hamilton2022unsupervised,melas2022deep}.
With these features, we take a gradual merging strategy to produce semantic-consistent region proposals at multiple granularities and save them efficiently with a tree structure.
In the training stage, given an image, we randomly select proposals from the corresponding merging tree in a top-down manner and further simulate possible clicks based on them, thus providing the model informative interactions for learning to understand interactions during the training process.

For training, the model is optimized by the provided interactions, which include meaningful regions and clicks.
Specifically, it takes the image and clicks as input and is optimized to produce similar segmentation as the region.
Moreover, we design a smoothness constraint when optimizing, for the purpose of mitigating the misleading brought by the inaccurate boundary of the region, which is caused by the down-sampling in the ViT.
The details of MIS and the training objective are described as follows.

\subsection{Multi-granularity Interaction Simulation}
\label{sec:proposal}

\noindent \textbf{Bottom-up Merging.}
At the core of our MIS is to discover semantic-consistent yet diverse regions, which is achieved efficiently by gradually merging.
In the beginning, we produce high-quality semantic features from a ViT~\cite{dosovitskiy2020image} trained in a self-supervised manner with DINO~\cite{caron2021emerging} to measure the semantic similarity and ensure the semantic-consistency of generated region proposals.
In the ViT, an image $ \mI \in \sR^{h \times w \times 3} $ is divided to non-overlapping $ n = \frac{h}{s} \times \frac{w}{s} $ patches with stride $ s $, which are then processed by some Transformer~\cite{vaswani2017attention} encoder layers.
The patch features $ \mF \in \sR^{n \times c} $ are obtained from the last layer of the ViT, where $ c $ is the feature dimension.
With the patch features, we implement the hierarchical merging process by Algorithm~\ref{algo:merging} and finally get a tree $ \bm{T} $ which records every merge process.

Initially, $ n $ patches form as $ n $ leaf nodes of the tree, and a set $ \sS $ is maintained for saving unmerged nodes, which initially contains all the leaf nodes.
In each iteration, the two unmerged nodes with the least cost according to their features will be merged to form a new node, until only one unmerged node remains.
Thus, there should be totally $ n - 1 $ iterations, and the index of the newly generated node in $ k $-th iteration is $ n + k $.
Moreover, considering the prior that parts belonging to the same object are usually adjacent, we add connectivity constraints when merging to greatly accelerate the search space for finding the minimum cost.
Specifically, each patch is only connected to its four adjacent patches and two regions are connected only if at least a pair of patches from them respectively are connected.

\begin{figure}[t]
    \centering
    \includegraphics[width=\linewidth]{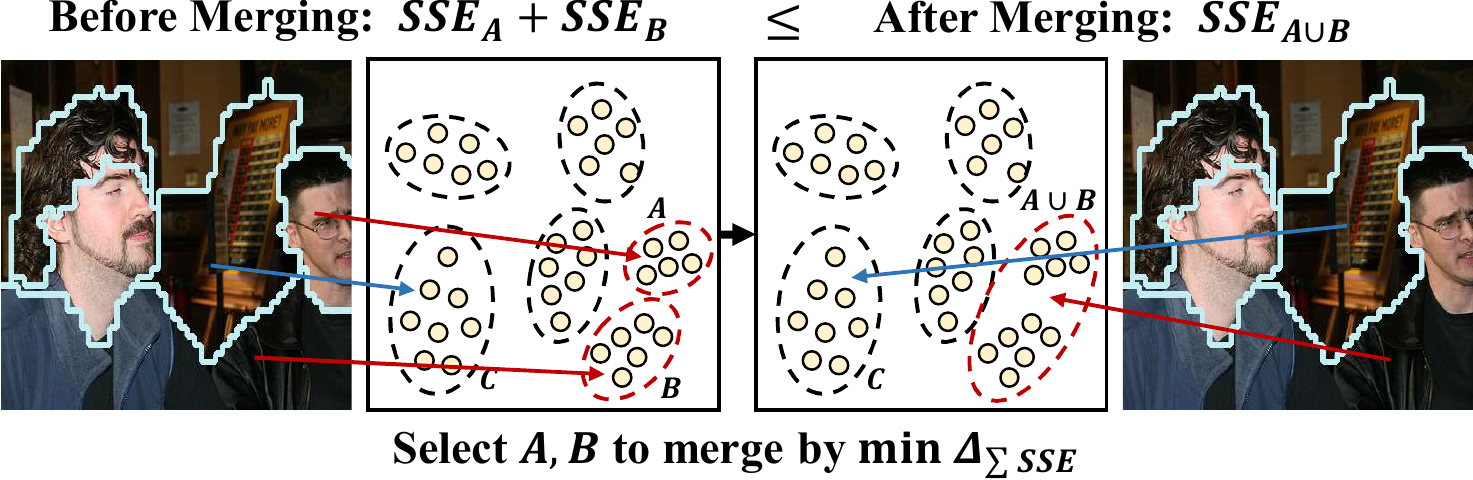}
    \caption{
        \textbf{Illustration of the cost of merging.}
        The points represent patch features and the circles denote regions that consist of the patches surrounded by it.
        Merging two regions into one leads to a more divergent cluster and the total SSE increases, thus the cost of merging can be measured as this increment.
    }
    \label{fig:ward}
\end{figure}

We follow Ward's method~\cite{ward1963hierarchical} to measure the cost of merging, which minimizes the increment in the total within-cluster sum of squared error (SSE) after merging, as shown in \Cref{fig:ward}.
Regarding a region $ \cA $ as a cluster that contains some patches, the SSE is defined as
\begin{equation}
    e_\cA = \sum_{i \in \cA} ||\mF_i - \vmu_\cA||^2,
\end{equation}
where $ \vmu_\cA \in \sR ^ c $ is the center of it, which can be computed as
\begin{equation}
    \vmu_\cA = \frac{1}{s_\cA} \sum_{i \in \cA} \mF_i,
\end{equation}
where $ s_\cA $ is the number of patches within $ \cA $.
When merging two regions, the new center is
\begin{equation}
    \vmu_{\cA \cup \cB} = \frac{s_\cA \vmu_\cA + s_\cB \vmu_\cB}{s_\cA + s_\cB}.
\label{eq:center}
\end{equation}
Then the cost of merging is defined as the difference of SSE before and after merging.
Since the SSE of all clusters except the clusters being merged are unchanged, it can be formulated as
\begin{equation}
\begin{aligned}
    \mathrm{Cost}(\cA, \cB) &= e_{\cA \cup \cB} - (e_\cA + e_\cB) \\
                &= \frac{s_\cA \cdot s_\cB}{s_\cA + s_\cB} ||\vmu_\cA - \vmu_\cB||^2.
\end{aligned}
\label{eq:cost}
\end{equation}
During the merging process, the center $ \vmu $ and the size $ s $ of each region are recorded.
Initially, the size is set to 1 for each patch and the center is its feature.
The detailed calculation after each merging is as follows:
\noindent \textbf{(1)} Computing the center of the newly generated region according to \Cref{eq:center}.
\noindent \textbf{(2)} Computing the size of the newly generated region as $ s_{\cA \cup \cB} = s_\cA + s_\cB $.
\noindent \textbf{(3)} Computing the cost of merging newly generated regions with other regions according to \Cref{eq:cost}.
Since the other regions are not changed, the cost between them remains the same.

\begin{algorithm}[t]
  \caption{Bottom-up Merging}
  \label{algo:merging}

  \SetKwInOut{Input}{Input}
  \SetKwInOut{Output}{Output}

  \Input{Patch features $ \mF \in \sR^{n \times c} $}
  \Output{Merge tree $ \mT \in \sZ^{(n - 1) \times 2} $}

  \emph{Initialization: $ \sS \leftarrow \{1,2,\dots,n\} $}

  \For{$ k \leftarrow 1 $ \KwTo $ n - 1 $}{
    $ i,j \leftarrow \underset{i,j \in \bm{S}}{\arg\min} \ \mathrm{Cost}(i,j,\mF) \ \mathrm{s.t.}\ \mathrm{Connect}(i,j) $\;
    $ \mT_{k, :} \leftarrow (i, j) $\;
    $  $
    $ \sS \leftarrow \complement_{\sS}{\{i, j\}} \cup \{n + k\} $\;
  }
\end{algorithm}

\begin{algorithm}[t]
  \caption{Top-down Sampling}
  \label{algo:sampling}

  \SetKwInOut{Input}{Input}
  \SetKwInOut{Output}{Output}

  \Input{Merge tree $ \mT $, decay coefficient $ \alpha $}
  \Output{Node index $ k $}

  \emph{Initialization: $ p \leftarrow 1.0, \ k \leftarrow n - 1 $}

  \While{$ \mathrm{Rand}() < p $ and $ k > 0 $}{
    \eIf{$ \mathrm{Rand}() < 0.5 $}{$ k \leftarrow \mT_{k,0} - n $\;}{$ k \leftarrow \mT_{k,1} - n $\;}
    $ p \leftarrow \alpha \cdot p $\;
  }
  
  $ k \leftarrow k + n $\;
\end{algorithm}

\begin{figure*}[t]
    \centering
    \includegraphics[width=\textwidth]{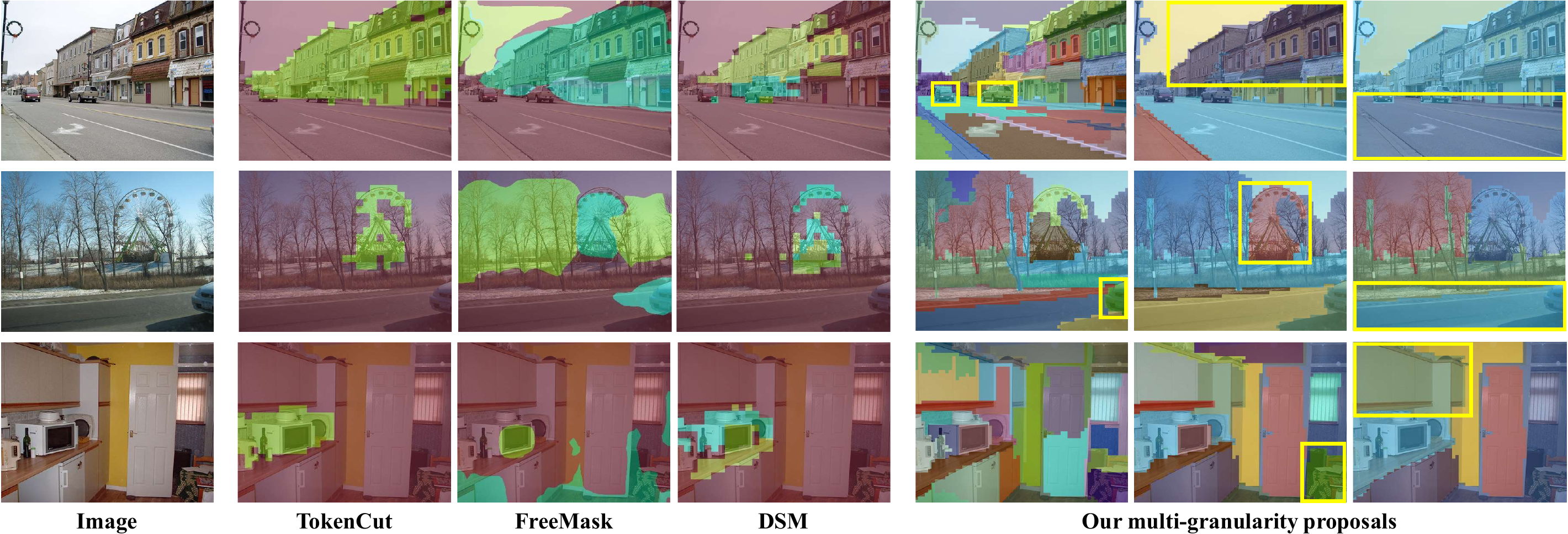}
    \caption{
        \textbf{Qualitative results of discovered regions for the baselines and our MIS.}
        While the baselines only focus on some salient regions in a fixed granularity, our MIS produces semantic-consistent regions at multiple granularities thus catching more objects and parts.  
    }
    \label{fig:proposal}
\end{figure*}

After $ n - 1 $ iterations, we get a tree structure that is stored by a matrix $ \mT \in \sZ^{(n - 1) \times 2} $, where the $ i $-th row is the indices of the two children of the $ i + n $ node.
In the tree, each node represents a merge and also represents a region proposal which consists of all patches within the subtree rooted at this node.
Hence, with this hierarchical relationship, we obtain $ n - 1 $ region proposals at multiple granularities for each image, which help the interactive segmentation model to learn various interactions.
For storage, there is no need to store all the $ n - 1 $ masks but only the tree structure, that is, the matrix $ \mT $, which makes it more space-efficient.

\smallskip

\noindent \textbf{Top-down Sampling.}
When training, the MIS is worked based on the stored merging tree.
Specifically, when sampling an image for training, we select a mask proposal from its corresponding merging tree for simulating an interaction. 
In the merging tree, shallow nodes correspond to relatively complete parts or their combinations and deeper nodes are more relevant to finely divided parts.
Intuitively, we prefer the complete one while maintaining diversity.
Therefore, we design a sampling strategy in a top-down manner and gradually decrease the probability to go deeper, as shown in Algorithm~\ref{algo:sampling}.
Concretely, we start at the root node (\emph{i.e.}, the node with an index of $ 2n - 1 $) and the probability $ p = 1.0 $.
Then in each iteration, we choose to go deeper or stop with probability $ p $, and decrease $ p $ with a coefficient $ \alpha $.
The loop will terminate when choosing to stop or reach a leaf node (\emph{i.e.}, nodes whose index is less than or equal to $ n $).
If it is decided to go deeper, we randomly move the current node to one of its children.
Finally, we get the index of a node $ k $ corresponding to a region proposal, where the patches in the sub-tree rooted at $ k $ are annotated as foreground.

\smallskip

\noindent \textbf{Click Sampling.}
With the sampled region, the last step is to simulate possible clicks according to the region.
Drawing on the experience of previous methods, we use a combination of random and iterative click simulation strategies to produce a click map following RITM~\cite{sofiiuk2022reviving}.
Specifically, some positive and negative clicks are first generated in parallel according to the foreground and background defined by the region.
Then some additional clicks are appended by checking the erroneous region of the prediction which is produced by previous clicks.
Finally, the MIS generates complete interactions which consist of a segmentation target and corresponding clicks.

\subsection{Training}

\noindent \textbf{Interactive Segmention Model.}
Typical interactive segmentation models are built based on semantic segmentation models (\emph{e.g.}, FCN~\cite{long2015fully}) by taking an additional click map as input, which is transformed from the coordinate of clicks.
Formulaically, given an image $ \mI \in \sR^{h \times w \times 3} $, a click map $ \mD \in \sR^{h \times w \times 2} $ which presents positive and negative clicks respectively, and previous prediction $ \mB \in \{0, 1\}^{h \times w} $, the model $ \mathcal{F}(\cdot) $ outputs the probability of being foreground for each pixel
\begin{equation}
    \mQ = \mathcal{F}(\mathrm{Fusion}(\mI, \mD, \mB)),\ \mQ \in [0,1]^{h \times w},
\end{equation}
where $ \mathrm{Fusion}(\cdot) $ is some kind of fusion operation such as concatenate.
Through training, the model learns to map the image and the user clicks to foreground and background segmentation thus can be used interactively when inference.

\smallskip

\begin{table*}[t]
  \footnotesize
  \centering
  \renewcommand{\arraystretch}{1.2}
  \setlength{\tabcolsep}{3.63pt}
  \begin{tabular*}{\textwidth}{l l | c c | c c | c c | c c}
    \thickhline

    \multirow{2}{*}{\textbf{Method}} & \multirow{2}{*}{\textbf{Backbone}} & 
    \multicolumn{2}{c|}{\textbf{GrabCut}} & \multicolumn{2}{c|}{\textbf{Berkeley}} & \multicolumn{2}{c|}{\textbf{SBD}} & \multicolumn{2}{c}{\textbf{DAVIS}} \\
    \cline{3-10}
    & & \textbf{NoC@85 $ \downarrow $} & \textbf{NoC@90 $ \downarrow $} & \textbf{NoC@85 $ \downarrow $} & \textbf{NoC@90 $ \downarrow $} & \textbf{NoC@85 $ \downarrow $} & \textbf{NoC@90 $ \downarrow $} & \textbf{NoC@85 $ \downarrow $} & \textbf{NoC@90 $ \downarrow $} \\

    \hline

    \rowcolor[gray]{0.9}
    \multicolumn{10}{l}{\emph{Deep Supervised Methods}} \\

    DIOS w/o GC~\cite{xu2016deep} \tiny{$_\text{CVPR16}$} & FCN
    & \underline{8.02} & \underline{12.59} & - & - & \underline{14.30} & \underline{16.79} & \underline{12.52} & \underline{17.11} \\
    DIOS w/ GC~\cite{xu2016deep} \tiny{$_\text{CVPR16}$} & FCN
    & \underline{5.08} & \underline{6.08} & - & - & \underline{9.22} & \underline{12.80} & \underline{9.03} & \underline{12.58} \\
    LD~\cite{li2018interactive} \tiny{$_\text{CVPR18}$} & VGG-19
    & \underline{3.20} & \underline{4.79} & - & - & \underline{7.41} & \underline{10.78} & 5.05 & \underline{9.57} \\
    BRS~\cite{jang2019interactive} \tiny{$_\text{CVPR19}$} & DenseNet
    & \underline{2.60} & \underline{3.60} & - & \underline{5.08} & 6.59 & \underline{9.78} & 5.58 & 8.24 \\
    f-BRS~\cite{sofiiuk2020f} \tiny{$_\text{CVPR20}$} & ResNet-101
    & \underline{2.30} & \underline{2.72} & - & 4.57 & 4.81 & 7.73 & 5.04 & 7.41 \\
    FCA-Net~\cite{lin2020interactive} \tiny{$_\text{CVPR20}$} & ResNet-101
    & - & 2.08 & - & 3.92 & - & - & - & 7.57 \\
    IA+SA~\cite{kontogianni2020continuous} \tiny{$_\text{ECCV20}$} & ConvNet
    & - & \underline{3.07} & - & \underline{4.94} & - & - & 5.16 & - \\
    CDNet~\cite{chen2021conditional} \tiny{$_\text{ICCV21}$} & ResNet-34
    & 1.86 & 2.18 & 1.95 & 3.27 & 5.18 & 7.89 & 5.00 & 6.89 \\ 
    RITM~\cite{sofiiuk2022reviving} \tiny{$_\text{ICIP22}$} & HRNet-18
    & 1.76 & 2.04 & 1.87 & 3.22 & 3.39 & 5.43 & 4.94 & 6.71 \\
    FocalClick~\cite{chen2022focalclick} \tiny{$_\text{CVPR22}$} & MiT-B0 
    & 1.66 & 1.90 & - & 3.14 & 4.34 & 6.51 & 5.02 & 7.06 \\
    FocusCut~\cite{lin2022focuscut} \tiny{$_\text{CVPR22}$} & ResNet-101
    & 1.46 & 1.64 & 1.81 & 3.01 & 3.40 & 5.31 & 4.85 & 6.22 \\
    PseudoClick~\cite{liu2022pseudoclick} \tiny{$_\text{ECCV22}$} & HRNet-32
    & - & 1.84 & - & 2.98 & - & 5.61 & 4.74 & 6.16 \\
    SimpleClick~\cite{liu2022simpleclick} \tiny{$_\text{arXiv22}$} & ViT-B 
    & \textbf{1.40} & \textbf{1.54} & \textbf{1.44} & \textbf{2.46} & \textbf{3.28} & \textbf{5.24} & \textbf{4.10} & \textbf{5.48} \\

    \hline

    \hline
    \hline

    \rowcolor[gray]{0.9}
    \multicolumn{10}{l}{\emph{Non-deep Learning Methods}} \\

    GraphCut~\cite{boykov2001interactive} \tiny{$_\text{ICCV01}$} & N/A
    & 7.98 & 10.00 & - & 14.22 & 13.60 & 15.96 & 15.13 & 17.41 \\
    Random Walk~\cite{grady2006random} \tiny{$_\text{TPAMI06}$} & N/A
    & 11.36 & 13.77 & - & 14.02 & 12.22 & 15.04 & 16.71 & 18.31 \\
    Geodesic Matting~\cite{bai2007geodesic} \tiny{$_\text{IJCV09}$} & N/A
    & 13.32 & 14.57 & - & 15.96 & 15.36 & 17.60 & 18.59 & 19.50 \\
    GSC~\cite{gulshan2010geodesic} \tiny{$_\text{CVPR10}$} & N/A
    & 7.10 & 9.12 & - & 12.57 & 12.69 & 15.31 & 15.35 & 17.52 \\
    ESC~\cite{gulshan2010geodesic} \tiny{$_\text{CVPR10}$} & N/A
    & 7.24 & 9.20 & - & 12.11 & 12.21 & 14.86 & 15.41 & 17.70 \\

    \rowcolor[gray]{0.9}
    \multicolumn{10}{l}{\emph{Deep Unsupervised Methods}} \\

    TokenCut$^\star$~\cite{wang2022self} {\tiny{$_\text{CVPR22}$}} & ViT-B
    & 3.04 & 5.74 & 6.30 & 9.97 & 10.82 & 13.16 & 10.56 & 15.01 \\
    FreeMask$^\star$~\cite{wang2022freesolo} {\tiny{$_\text{CVPR22}$}} & ViT-B
    & 4.06 & 6.10 & 5.55 & 9.02 & 8.61 & 12.01 & 8.26 & 13.05 \\
    DSM$^\star$~\cite{melas2022deep} {\tiny{$_\text{CVPR22}$}} & ViT-B
    & 3.64 & 4.64 & 5.49 & 7.75 & 8.59 & 11.57 & 7.08 & 10.11 \\

    \hline

    \rowcolor{aliceblue!80}
    MIS w/o $ \mathcal{L}_{smooth} $ (\emph{Ours}) & ViT-B 
    & 2.08 & 2.56 & 3.40 & 5.51 & 7.38 & 9.92 & 6.78 & 10.08 \\

    \rowcolor{aliceblue!80}
    MIS (\emph{Ours}) & ViT-B 
    & \textbf{1.94} & \textbf{2.32} & \textbf{3.09} & \textbf{4.58} & \textbf{6.91} & \textbf{9.51} & \textbf{6.33} & \textbf{8.44} \\
    
    \thickhline
  \end{tabular*}
  \vspace{1pt}
  \caption{
        \textbf{Comparison with previous methods and baselines on four interactive segmentation benchmarks.}
        $ \star $ denotes baselines that we introduce from other unsupervised tasks.
        From top to bottom we show the previous deep supervised method, unsupervised non-deep learning method, and our newly proposed deep unsupervised method.
        All models listed here are trained on SBD dataset.
        Our MIS achieves superior performance within unsupervised methods and baselines and even exceeds early deep supervised methods (denoted by \underline{underline}).
    }
  \label{tab:main}
\end{table*}

\smallskip

\noindent \textbf{Training with Proposals.}
Under our unsupervised setting, the model learns this mapping by the simulated interactions from the MIS.
The model is first learned by the supervision from the multi-granularity region proposals $ \mM $ by optimizing the objective
\begin{equation}
    \mathcal{L}_{pseudo} = \mathrm{BCE}(\mM, \mQ),
\end{equation}
where $ \mathrm{BCE}(\cdot) $ denotes the binary cross entropy loss.
Although the region proposals generated by the MIS contains semantics, the possible inaccuracy of the self-supervised feature extractor and the patching process in the ViT will bring some inaccurate part.
Fortunately, as discovered by previous research~\cite{arpit2017closer} that the model tends to learn effect patterns rather than memorize the noise, we find that the model can learn to correctly map clicks to segmentation after training with these simulated interactions.

\smallskip

\noindent \textbf{Smoothness Constraint.}
In order to help the model correct the error caused by patching, we propose a smoothness constraint.
We first introduce the bilateral affinity matrix~\cite{tomasi1998bilateral,barron2016fast} $ \mW $ which presents the affinity of pixels based on their low-level feature and the item $ \mW_{i,j} $ is
\begin{equation}
    exp(-\frac{||\va_i^{xy} - \va_j^{xy}||^2}{2\sigma_{xy}^2} - \frac{(a_i^l - a_j^l)^2}{2\sigma_l^2} - \frac{||\va_i^{uv} - \va_j^{uv}||^2}{2\sigma_{uv}^2}), \nonumber
\end{equation}
where $ \va_i $ is a pixel with a spatial position $ (p_i^x, p_i^y) $ and color $ (p_i^l, p_i^u, p_i^v) $ and $ \sigma_{xy}, \sigma_l, \sigma_{uv} $ control the extent of the spatial, luma, and chroma support of the filter.
Motivated by the bilateral solver~\cite{barron2016fast}, the smooth constraint is implemented by restricting the model to produce consistent predictions on pixels with high bilateral affinity, which can be formulated as
\begin{equation}
    \mathcal{L}_{smooth} = \sum_i \frac{1}{|\mathcal{N}_i|} \sum_{j \in \mathcal{N}_i} \mW_{ij} \cdot (\mQ_i - \mQ_j)^2.
\end{equation}
As we mainly concern with the predictions on the boundary, we only perform the loss locally for fast computation and memory efficiency, \emph{i.e.}, for pixel $ i $ the loss is calculated in its $ 5 \times 5 $ neighborhoods $ \mathcal{N}_i $.
Finally, the overall training objective is
\begin{equation}
\label{eq:loss}
    \mathcal{L} = \mathcal{L}_{pseudo} + \lambda \cdot \mathcal{L}_{smooth},
\end{equation}
where $ \lambda $ controls the strength of the smoothness constraint.

\begin{figure*}[t]
    \centering
    \includegraphics[width=\textwidth]{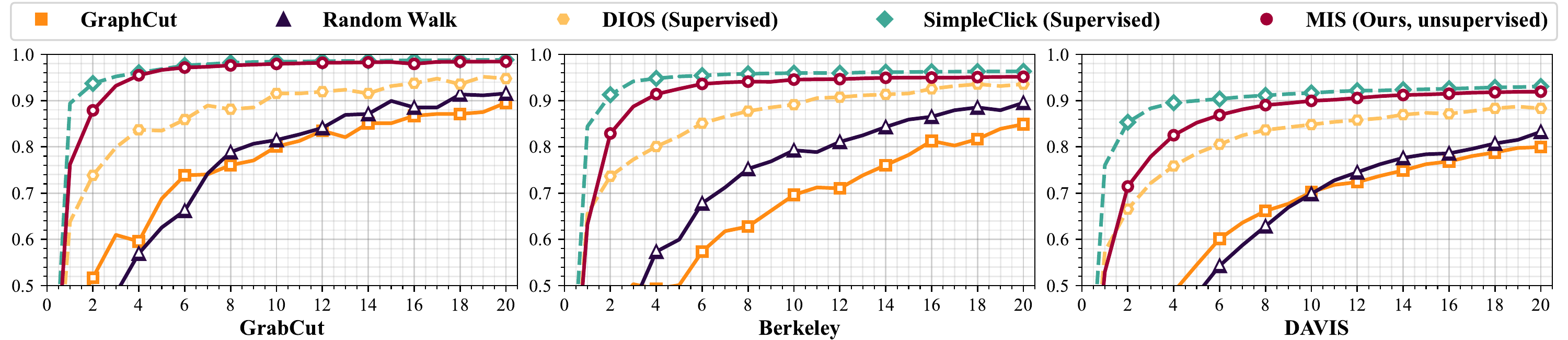}
    \caption{\textbf{Comparison of IoU-NoC curve.} The two dotted lines indicate the performance of supervised approaches. The proposed method achieves surprising performance that significantly exceeds traditional methods and is close to supervised method when the NoC increases.}
    \label{fig:curve}
\end{figure*}

\begin{table*}[t]
  \footnotesize
  \centering
  \renewcommand{\arraystretch}{1.2}
  \setlength{\tabcolsep}{8.08pt}
  \begin{tabular*}{\textwidth}{c | l l | l l | l l | l l}
    \thickhline

     \multirow{2}{*}{\textbf{Value}} & \multicolumn{2}{c|}{\textbf{GrabCut}} & \multicolumn{2}{c|}{\textbf{Berkeley}} & \multicolumn{2}{c|}{\textbf{SBD}} & \multicolumn{2}{c}{\textbf{DAVIS}} \\
    \cline{2-9}
    & \textbf{NoC@85 $ \downarrow $} & \textbf{NoC@90 $ \downarrow $} & \textbf{NoC@85 $ \downarrow $} & \textbf{NoC@90 $ \downarrow $} & \textbf{NoC@85 $ \downarrow $} & \textbf{NoC@90 $ \downarrow $} & \textbf{NoC@85 $ \downarrow $} & \textbf{NoC@90 $ \downarrow $} \\

    \hline
    \rowcolor[gray]{0.9} \multicolumn{9}{l}{\emph{Decay coefficient $ \alpha $}} \\

    0.0 & 2.12 & 2.48 & 3.89 & 5.52 & 8.71 & 10.87 & 7.84 & 9.92 \\

    0.8 & \textbf{1.94} \decrease{0.18} & 2.40 \decrease{0.08} & \textbf{3.08} \decrease{0.81} & 4.99 \decrease{0.53} & 7.03 \decrease{1.68} & 9.62 \decrease{1.25} & 6.36 \decrease{1.48} & 8.57 \decrease{1.35} \\

    0.9 & \textbf{1.94} \decrease{0.18} & \textbf{2.32} \decrease{0.16} & 3.09 \decrease{0.80} & \textbf{4.58} \decrease{0.94} & \textbf{6.91} \decrease{1.80} & \textbf{9.51} \decrease{1.36} & 6.33 \decrease{1.51} & \textbf{8.44} \decrease{1.48} \\

    0.95 & 2.22 \increase{0.10} & 2.62 \increase{0.14} & 3.19 \decrease{0.70} & 4.89 \decrease{0.63} & \textbf{6.91} \decrease{1.80} & 9.52 \decrease{1.35} & \textbf{6.09} \decrease{1.75} & 8.54 \decrease{1.38} \\

    \hline
    \rowcolor[gray]{0.9} \multicolumn{9}{l}{\emph{Strength of the Smoothness Constraint $ \lambda $}} \\

    0.0 & 2.08 & 2.56 & 3.40 & 5.51 & 7.38 & 9.92 & 6.78 & 10.08 \\

    3.0 & 2.02 \decrease{0.06} & 2.50 \decrease{0.06} & 3.07 \decrease{0.33} & 5.04 \decrease{0.47} & 6.84 \decrease{0.54} & \textbf{9.33} \decrease{0.59} & \textbf{6.32} \decrease{0.46} & 8.78 \decrease{1.30} \\

    5.0 & \textbf{1.94} \decrease{0.14} & 2.34 \decrease{0.22} & \textbf{3.01} \decrease{0.39} & 4.71 \decrease{0.80} & \textbf{6.79} \decrease{0.59} & 9.39 \decrease{0.53} & 6.39 \decrease{0.39} & 8.55 \decrease{1.53} \\

    10.0 & \textbf{1.94} \decrease{0.14} & \textbf{2.32} \decrease{0.24} & 3.09 \decrease{0.31} & \textbf{4.58} \decrease{0.93} & 6.91 \decrease{0.47} & 9.51 \decrease{0.41} & 6.33 \decrease{0.45} & \textbf{8.44} \decrease{1.64} \\

    15.0 & 2.12 \increase{0.04} & 2.50 \decrease{0.06} & 3.25 \decrease{0.15} & 5.13 \decrease{0.38} & 7.12 \decrease{0.26} & 9.85 \decrease{0.07} & 6.71 \decrease{0.07} & 9.10 \decrease{0.98} \\

    \thickhline
  \end{tabular*}
  \vspace{1pt}
  \caption{
    \textbf{Ablation on multi-granularity proposals and smoothness constraint.}
    All results show their effectiveness and robustness.
   }
  \label{tab:ablation}
\end{table*}

\section{Experiments}

\noindent \textbf{Implementation Details.}
We follow SimpleClick~\cite{liu2022simpleclick} to build the interactive segmentation model, which consists of two patch embedding modules for image and click map respectively, a ViT~\cite{dosovitskiy2020image} backbone initialized with MAE~\cite{he2022masked}, a simple feature pyramid~\cite{li2022exploring}, and an MLP segmentation head.
For training, we optimize the model for 55 epochs using the Adam~\cite{kingma2014adam} optimizer with a learning rate of 5e-5, which decays by a factor of 10 at 50 epochs.
The decay coefficient $ \alpha $ in Algorithm~\ref{algo:sampling} and the weight $ \lambda $ in \Cref{eq:loss} are 0.9 and 10.0, respectively.
In addition, we set a threshold of 0.05 to filter small proposals.
That is to say, we ignore the proposals whose areas are less than 0.05 times the image area.
For inference, we set the threshold of binarizing the prediction to 0.5 as normal binary segmentation and use the same pre-processing as \cite{liu2022simpleclick}.
To ensure the fully unsupervised setting, we employ a self-supervisedly pre-trained model as the feature extractor.
Specifically, a Vision Transformer (ViT)~\cite{dosovitskiy2020image} trained with DINO~\cite{caron2021emerging} is adopted since the property of representing patch-level semantics~\cite{caron2021emerging,hamilton2022unsupervised,simeoni2021localizing}.
In order to prevent the extracted features from ignoring small objects or parts, we use ViT-Small~\cite{simeoni2021localizing} with a patch size of 8.
Before feeding an image into the ViT, we resize the image to make its height and width divisible by the patch size (\emph{i.e.}, the target height is $ ( h + p - h \mod p ) $ when the original height $ h $ is not divisible by the patch size $ p $), and the position embedding of the ViT is interpolated to fit the image size using bilinear interpolation.
We finally use all output tokens of the last block except the \emph{cls} token as the patch features.

\smallskip

\noindent \textbf{Evaluation.}
The evaluation is done under the standard evaluation protocol of click-based interactive segmentation.
Specifically, we adopt the same click simulator as previous work~\cite{xu2016deep,jang2019interactive,lin2020interactive,sofiiuk2020f,chen2021conditional,sofiiuk2022reviving,lin2022focuscut,chen2022focalclick} to sample clicks when evaluation.
Roughly, the next click will be placed at the center of the largest error region by comparing the ground-truth and prediction.
We adopt the Number of Click (NoC) as the evaluation metric, which counts the average number of clicks needed to achieve a fixed Intersection over Union (IoU), the smaller the better.
We use two commonly used target intersection-over-union (IoU) thresholds 85\% and 90\%, denoted as NoC@85 and NoC@90 respectively.
Additionally, the IoU-NoC curves are also adopted to represent the convergence trend of IoU when the NoC increases, as well as compare the IoU metric under the same NoC.

\smallskip

\noindent \textbf{Dataset.}
We evaluate the performance of our MIS on the following four benchmarks.
(1) \textbf{GradCut}~\cite{rother2004grabcut}: The dataset contains 50 relatively easy images whose background and foreground have a clear difference.
(2) \textbf{Berkeley}~\cite{mcguinness2010comparative}: The dataset contains 96 images with 100 instances and some of them are more challenging than GrabCut.
(3) \textbf{SBD}~\cite{hariharan2011semantic}: The dataset contains 2,857 images with 6,671 challenging instances for evaluation, which does not overlap with our training set.
(4) \textbf{DAVIS}~\cite{perazzi2016benchmark}: The dataset contains 50 high-quality videos. We follow previous works~\cite{lin2022focuscut,chen2022focalclick,liu2022pseudoclick,sofiiuk2022reviving,liu2022simpleclick} and use the same 345 frames for evaluation.

\begin{figure*}[t]
    \centering
    \includegraphics[width=\textwidth]{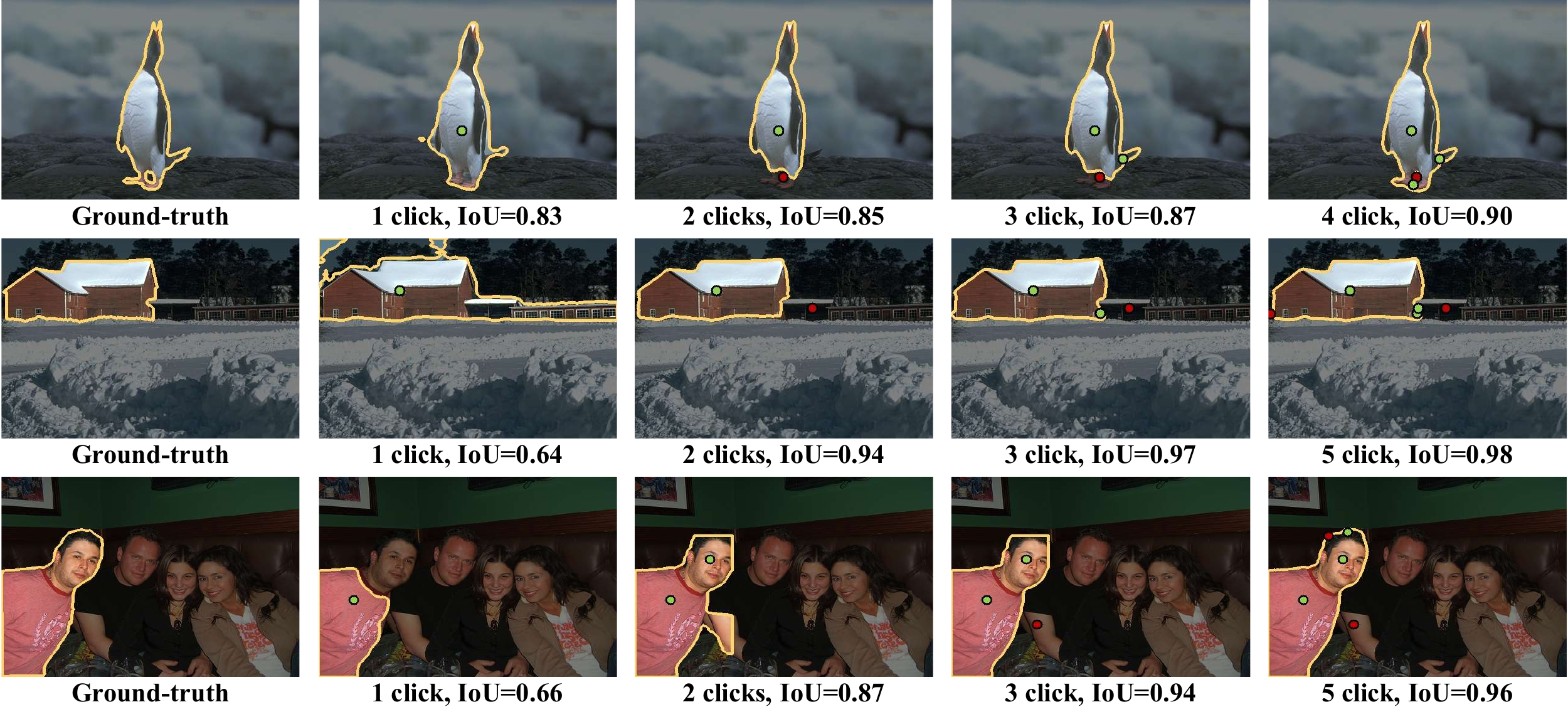}
    \caption{
        \textbf{Visualization of the interaction and segmentation.}
        The green and red points indicate positive and negative clicks respectively.
    }
    \label{fig:visualization}
\end{figure*}

\begin{table*}[t]
  \footnotesize
  \centering
  \renewcommand{\arraystretch}{1.25}
  \setlength{\tabcolsep}{3.15pt}
  \begin{tabular*}{\textwidth}{c l | l l | l l | l l | l l}
    \thickhline

    \multirow{2}{*}{\textbf{Data Ratio}} & \multirow{2}{*}{\textbf{Initialization}} & 
    \multicolumn{2}{c|}{\textbf{GrabCut}} & \multicolumn{2}{c|}{\textbf{Berkeley}} & \multicolumn{2}{c|}{\textbf{SBD}} & \multicolumn{2}{c}{\textbf{DAVIS}} \\
    \cline{3-10}
    & & \textbf{NoC@85 $ \downarrow $} & \textbf{NoC@90 $ \downarrow $} & \textbf{NoC@85 $ \downarrow $} & \textbf{NoC@90 $ \downarrow $} & \textbf{NoC@85 $ \downarrow $} & \textbf{NoC@90 $ \downarrow $} & \textbf{NoC@85 $ \downarrow $} & \textbf{NoC@90 $ \downarrow $} \\

    \hline

    \multirow{2}{*}{1\%} & Default (MAE~\cite{he2022masked}) & 2.32 & 2.52 & 2.36 & 4.44 & 8.01 & 10.43 & 5.81 & 8.28 \\
    & MIS (\emph{Ours}) & \textbf{1.54} \decrease{0.78} & \textbf{1.90} \decrease{0.62} & \textbf{1.95} \decrease{0.41} & \textbf{3.73} \decrease{0.71} & \textbf{5.57} \decrease{2.44} & \textbf{8.15} \decrease{2.28} & \textbf{5.36} \decrease{0.45} & \textbf{7.40} \decrease{0.88} \\

    \rowcolor[gray]{0.9}
    & Default (MAE~\cite{he2022masked}) & 1.68 & 2.22 & 1.97 & 3.64 & 5.48 & 8.13 & 5.14 & 7.23 \\
    \rowcolor[gray]{0.9}
    \multirow{-2}{*}{5\%} & (\emph{Ours}) & \textbf{1.62} \decrease{0.06} & \textbf{1.88} \decrease{0.34} & \textbf{1.95} \decrease{0.02} & \textbf{3.32} \decrease{0.32} & \textbf{5.21} \decrease{0.27} & \textbf{7.49} \decrease{0.64} & \textbf{4.54} \decrease{0.60} & \textbf{6.11} \decrease{1.12} \\

    \multirow{2}{*}{10\%} & Default (MAE~\cite{he2022masked}) & 1.64 & 1.84 & 1.88 & 3.21 & 5.33 & 7.89 & 4.88 & 6.77 \\
    & MIS (\emph{Ours}) & \textbf{1.40} \decrease{0.24} & \textbf{1.50} \decrease{0.34} & \textbf{1.64} \decrease{0.24} & \textbf{2.73} \decrease{0.48} & \textbf{4.61} \decrease{0.72} & \textbf{6.95} \decrease{0.94} & \textbf{4.25} \decrease{0.63} & \textbf{5.94} \decrease{0.83} \\

    \thickhline
  \end{tabular*}
  \vspace{1pt}
  \caption{
    \textbf{Performance with limited annotations.} When training the model in a supervised manner with a limited number of annotations (1\%, 5\%, 10\%), the performance can be significantly improved by unsupervised pre-training using the proposed method.}
  \label{tab:few}
\end{table*}

\subsection{Quantitative Results}
\label{sec:results}

\noindent \textbf{Baselines.}
Since the fully unsupervised interactive segmentation has not been explored by previous research, we introduce some baselines which are originally designed for unsupervised salient detection (TokenCut~\cite{wang2022self}), instance segmentation (FreeMask~\cite{wang2022freesolo}), and semantic segmentation (DSM~\cite{melas2022deep}) to demonstrate the effectiveness of our method.
We follow the official implementation of these baselines to extract mask proposals and use them to train the same model as ours.
We denote these baselines by ``$ \star $" in \Cref{tab:main}.

\smallskip

\noindent \textbf{Comparison with Previous Work.}
For comparison, we present experimental results of previous supervised deep learning methods, non-deep learning methods, baselines, and the proposed MIS on four aforementioned datasets.
Under the unsupervised setting, our method significantly outperforms all other methods as shown in the second part of \Cref{tab:main}.
It is also worth mentioning that the proposed MIS achieves inspiring performance that is even comparable with some previous supervised approaches.
This support that the interactive segmentation task can be solved in a more label-efficient way.
Moreover, we show the IoU-NoC curve of previous interactive segmentation methods and MIS in \Cref{fig:curve}, where the dotted lines indicate the upper and lower performance of previous supervised methods.
It can be found that our method reaches obviously higher IoU compared to the non-deep learning methods, especially in the first few clicks, and gradually converge to the supervised counterpart when NoC increases.

\smallskip

\noindent \textbf{Comparison with Baselines.}
MIS also gets advanced results compared to the baselines which only focus on the specific granularity such as the most salient object or fixed number of objects, demonstrating the effectiveness of diverse interactions under the unsupervised setting.

\smallskip

\noindent \textbf{Supervised Fine-tuning.}
Besides the fully unsupervised setting, we further fine-tune our model in a supervised manner with only a limited number of annotations~(1\%, 5\%, 10\%).
As shown in \Cref{tab:few}, the model initialized with MIS pre-training
the MAE~\cite{he2022masked} counterpart especially on the relatively hard dataset (\emph{i.e.}, SBD and DAVIS).
These results demonstrate that MIS can serve as a strong unsupervised pre-training method for interactive segmentation when the available annotations are rare.

\begin{table*}[t]
  \footnotesize
  \centering
  \renewcommand{\arraystretch}{1.2}
  \setlength{\tabcolsep}{3.58pt}
  \begin{tabular*}{\textwidth}{c | l | l l | l l | l l | l l}
    \thickhline

    \multirow{2}{*}{\textbf{Connectivity}} & \multirow{2}{*}{\makecell[c]{\textbf{Speed} \\ \textbf{(image/s) $ \uparrow $}}} & \multicolumn{2}{c|}{\textbf{GrabCut}} & \multicolumn{2}{c|}{\textbf{Berkeley}} & \multicolumn{2}{c|}{\textbf{SBD}} & \multicolumn{2}{c}{\textbf{DAVIS}} \\
    \cline{3-10}
    & & \textbf{NoC@85 $ \downarrow $} & \textbf{NoC@90 $ \downarrow $} & \textbf{NoC@85 $ \downarrow $} & \textbf{NoC@90 $ \downarrow $} & \textbf{NoC@85 $ \downarrow $} & \textbf{NoC@90 $ \downarrow $} & \textbf{NoC@85 $ \downarrow $} & \textbf{NoC@90 $ \downarrow $} \\

    \hline

    \ding{56} & 1.10 & 2.12 & 2.52 & \textbf{3.08} & 4.69 & \textbf{6.88} & \textbf{9.46} & \textbf{5.63} & \textbf{7.56} \\

    \rowcolor[gray]{0.9} \ding{51} & \textbf{3.73} \increase{3.39 $ \times $} & \textbf{1.94} \decrease{0.18} & \textbf{2.32} \decrease{0.20} & 3.09 \increase{0.01} & \textbf{4.58} \decrease{0.11} & 6.91 \increase{0.03} & 9.51 \increase{1.36} & 6.33 \increase{0.70} & 8.44 \increase{0.88} \\

    \thickhline
  \end{tabular*}
  \vspace{1pt}
  \caption{\textbf{Ablation on connectivity constraint.} The performance of using or not using connectivity constraint is not much different, but using connectivity constraints greatly improves the speed (3.39 $ \times $ improvement) and make the pre-processing more efficient.}
  \label{tab:connectivity}
  \vspace{5pt}
\end{table*}

\begin{table*}[ht]
  \footnotesize
  \centering
  \renewcommand{\arraystretch}{1.2}
  \setlength{\tabcolsep}{7.34pt}
  \begin{tabular*}{\textwidth}{c | l l | l l | l l | l l}
    \thickhline

    \multirow{2}{*}{\textbf{Sampling}} & \multicolumn{2}{c|}{\textbf{GrabCut}} & \multicolumn{2}{c|}{\textbf{Berkeley}} & \multicolumn{2}{c|}{\textbf{SBD}} & \multicolumn{2}{c}{\textbf{DAVIS}} \\
    \cline{2-9}
    & \textbf{NoC@85 $ \downarrow $} & \textbf{NoC@90 $ \downarrow $} & \textbf{NoC@85 $ \downarrow $} & \textbf{NoC@90 $ \downarrow $} & \textbf{NoC@85 $ \downarrow $} & \textbf{NoC@90 $ \downarrow $} & \textbf{NoC@85 $ \downarrow $} & \textbf{NoC@90 $ \downarrow $} \\

    \hline

    Uniform & 3.84 & 4.32 & 4.42 & 6.20 & 7.11 & 9.95 & 7.71 & 10.10 \\

    \rowcolor[gray]{0.9} Top-down & \textbf{1.94} \decrease{1.90} & \textbf{2.32} \decrease{2.00} & \textbf{3.09} \decrease{1.33} & \textbf{4.58} \decrease{1.62} & \textbf{6.91} \decrease{0.20} & \textbf{9.51} \decrease{0.44} & \textbf{6.33} \decrease{1.38} & \textbf{8.44} \decrease{1.66} \\

    \thickhline
  \end{tabular*}
  \vspace{1pt}
  \caption{\textbf{Ablation on sampling strategy.} The proposed top-down sampling strategy is effective because it reasonably assigns different weights to fragmented and complete components, so as to sample complete objects as much as possible while maintaining diversity.}
  \label{tab:sampling}
\end{table*}

\subsection{Qualitative Results}

\noindent \textbf{Multi-granularity Region Proposal.}
In \Cref{fig:proposal} we show the qualitative results of the region proposals generated by the baselines and our MIS, where different colors mean different mask proposals and our mask proposals.
Compared with them, our multi-granularity proposals capture more informative regions and are more diverse.
Through the top-down random sampling strategy, our proposals provide the model with rich interactions, which makes the model learn a better mapping from clicks to segmentation under the unsupervised setting.

\smallskip \noindent \textbf{Interactive Segmentation.}
\Cref{fig:visualization} show some interactive segmentation examples of our method.
The trained interactive segmentation model is able to capture the outline of the needed object by the first one or two clicks and achieves high IoU with a few clicks, which demonstrates the feasibility of our newly proposed unsupervised setting and the effectiveness of the MIS.
For some simple objects, our model can produce satisfactory predictions within the first few clicks.
It can also handle complex scenes such as partially occluded objects and adjacent instances.

\subsection{Ablation Study}

\noindent \textbf{Decay Coefficient.}
For the decay coefficient $ \alpha $, we take 0.0 as a baseline.
In this case, we always use one of the two children of the root node in the merging tree, thus multi-granularity is not allowed.
Comparing the results with $ \alpha = 0.8, 0.9, 0.95 $ we can find that the introduction of multi-granularity brings noticeable improvement, especially in the hard dataset.
It is also obvious that this hyper-parameter is relatively robust.

\smallskip

\noindent \textbf{Smoothness Constraint.}
For the smoothness constraint, we switch the weight $ \lambda $ in \Cref{eq:loss} to show its effectiveness.
The experimental results in \Cref{tab:ablation} demonstrate that the smoothness constraint is helpful for improving the performance through the principle that helps the model correct the boundary errors in the proposals.
The effectiveness is robust when $ \lambda $ is within a reason range.

\smallskip

\noindent \textbf{Connectivity.}
In bottom-up merging (Algorithm~\ref{algo:merging}), we introduce the connectivity constraint to restrict merges to only occur locally.
Here we discuss the effectiveness of it by ablation.
As shown in \Cref{tab:connectivity}, the performance of using or not using connectivity constraint is not much different, but using connectivity constraints greatly improves the speed (3.39 $ \times $ improvement) and make the pre-processing more efficient.
This is because we reduce the search space for the minimum cost from each pair of patches or regions to the adjacent patches or regions based on the prior on images.

\smallskip

\noindent \textbf{Top-down Sampling.}
The top-down sampling (Algorithm~\ref{algo:sampling}) is the core part of our MIS to achieve diversity.
We evaluate its effectiveness by comparing with a fully random strategy that randomly sample nodes on the merging tree from a uniform distribution.
The results in \Cref{tab:sampling} show clear advantages of our top-down sampling strategy.
When sampling uniformly, the finely divided object components are more likely to be selected due to the large number, resulting in local perception and thus requiring more clicks to localize objects.
In contrast, our top-down sampling strategy reasonably assigns different weights to fragmented and complete components, so as to sample complete objects as much as possible while maintaining diversity.

\section{Conclusion}

In this work, we open up a promising direction for deep interactive segmentation that completely discards the dependence on manual annotations.
While previous methods learn the interaction through object-oriented simulation with pixel-level annotations, we propose MIS and demonstrate the feasibility to simulate informative interaction with semantic-consistent regions discovered in an unsupervised manner.
Inspiring experimental results show the effectiveness of our method and the potential to reduce the labor of creating annotation without cost via training an interactive segmentation model unsupervisedly. 
Our experiment also supports that the interactive segmentation task can be solved in a more label-efficient way to trade off the performance and annotations.
We hope our exploration can promote the attendance of label efficiency on interactive segmentation.

{\small
\bibliographystyle{ieee_fullname}
\bibliography{main}
}

\end{document}